# Value Elimination: Bayesian Inference via Backtracking Search*

**Fahiem Bacchus**
Department of Computer Science
University of Toronto
Toronto, Ontario, Canada

**Shannon Dalmao**
Department of Computer Science
University of Toronto
Toronto, Ontario, Canada

**Toniann Pitassi**
Department of Computer Science
University of Toronto
Toronto, Ontario, Canada

## Abstract

We present **Value Elimination**, a new algorithm for Bayesian Inference. Given the same variable ordering information, Value Elimination can achieve performance that is within a constant factor of variable elimination or recursive conditioning, and on some problems it can perform exponentially better, *irrespective* of the variable ordering used by these algorithms. Value Elimination's other features include: (1) it can achieve the same space-time tradeoff guarantees as recursive conditioning; (2) it can utilize all of the logical reasoning techniques used in state of the art SAT solvers; these techniques allow it to obtain considerable extra mileage out of zero entries in the CPTs; (3) it can be naturally and easily extended to take advantage of context specific structure; and (4) it supports dynamic variable orderings which might be particularly advantageous in the presence of context specific structure. We have implemented a version of Value Elimination that demonstrates very promising performance, often being one or two orders of magnitude faster than a commercial Bayes inference engine, despite the fact that it does not as yet take advantage of context specific structure.

## 1 Introduction

Value elimination is a new algorithm for inference in Bayesian networks (BAYES). It represents an advance over previous algorithms in the sense that it can achieve all of their performance guarantees (up to constant factor), can provable achieve an exponential speedup on some problems, and, with some extra (polynomial) cost, can be considerably more flexibility than previous algorithms in terms of its ability to exploit context specific structure, logical reasoning, and more flexible variable orderings.

Value elimination is based on the algorithmic paradigm of backtracking, and was motivated by the close relationship between eliminating variables, as done in variable elimination, and instantiating variables, as done in backtracking algorithms.

Most popular algorithms for inference in Bayesian networks (BAYES) are based on the idea of eliminating variables. Starting with the initial conditional probability tables (CPTs) of the Bayes network, viewed as functions over local collections of the network variables, variable elimination involves summing out individual variables, in the process creating new functions over typically larger sets of variables.

Variable elimination can be used to solve a number of other problems (Dechter 1999). It has a close relationship to backtracking that is most apparent when we examine its application to SAT. SAT is the problem of determining whether or not a satisfying assignment exists for a CNF formula. The earliest algorithm for solving SAT was the Davis-Putnam procedure (DP) (Davis & Putnam 1960) which utilizes ordered resolution. This procedure is precisely variable elimination run on a symbolic representation of the local functions (each function is represented as a set of clauses). At each stage DP eliminates a variable, representing the new function thus created with a new set of clauses (Rish & Dechter 2000).

SAT can also be solved with a backtracking search procedure called DPLL (Davis, Logemann, & Loveland 1962). It turns out that in practice DPLL is *vastly superior* to DP. For example, in experimental data from 23 different SAT solvers (available on-line at the SatEx site (Simon & Chatalic 2001)), a version of DP that utilizes modern heuristics for computing good elimination orders (Dechter & Rish 1994) ranked last in its ability to solve problems. Its behavior on the "jhn" family of problems is typical. This family contains 50 problems each with 100 variables, 34 of which are unsatisfiable. The fastest DPLL based procedure is able to solve all 50 problems in 0.86 CPU seconds. The variable elimination DP algorithm was unable to solve any of these problems: on each problem it either runs out of memory, or exceeds a 10,000 CPU seconds time bound.

There are a number of reasons why DPLL can perform so much better. First, because it works at the level of assign-ments to variables (values) it can take advantage of *context specific structure* (i.e., structure that appears only after some set of variables have been assigned particular values). Second, it can utilize constraint propagation (e.g., unit propagation) to prune away much of its search space. And third, it can take further advantage of context specific structure through the use of dynamic variable orderings that can instantiate the variables in a different order along each branch of the search tree. By instantiating variables DPLL generates distinct subproblems, one for each value of the variable. It is free to solve each of these subproblems in a different way, which can be very advantageous as these subproblems are often structurally quite different. On the other hand, DP, and variable elimination in general, is always working with a single undifferentiated problem: eliminating a variable does not split the problem into subproblems. DP cannot treat different parts of the new problem in different ways, it must choose a single next variable to eliminate.

---

*This research funded by governments of Ontario and Canada through their NSERC and PREA programs.

These problem also exist in the popular jointree BAYES algorithms. These algorithms utilize a tree clustering that is based on a fixed elimination ordering, and they work at the level of variables rather than values. Although some techniques have been suggested for utilizing context specific structure, e.g., (Boutilier *et al.* 1996), these techniques still have to be retro-fitted into an algorithm that does not naturally accommodate such information. As a result not all such information can be exploited.

In this paper show that backtracking search can be profitably applied to reasoning in Bayes networks by developing a backtracking based BAYES algorithm we call Value Elimination. Some work on using backtracking search for various probabilistic reasoning tasks has already been done, e.g., (Walsh 2002; Poole 1996; Majercik & Littman 1998). Value Elimination, however, is based on a new generalization of dependency directed backtracking techniques to the probabilistic context. This generalization allows us to compute and cache information during the search so that redundant computations are eliminated, while still retaining the flexibility of using dynamic variable orderings and context specific structure. These techniques allows Value Elimination to achieve the same worst case performance guarantees as the current best algorithms for BAYES, and at the same time provably outperform them on some problems.

We also present an implementation of Value Elimination along with empirical evidence to show that the approach can in many cases be competitive with current BAYES algorithms. This is especially significant as our current implementation does not yet utilize context specific structure. Hence, the fact that it is already providing good performance is very encouraging.

In the rest of the paper we will first present a development of Value Elimination, pointing out how it can take advantage of context specific structure. We then show that it can simulate both variable elimination as well as recursive conditioning within a constant factor, thus showing that it achieves the same performance guarantees as these algorithms. We also show that there exists a class of problems on which it can perform exponentially better. Finally, we present some encouraging empirical results from our implementation, and give some conclusions.

## 2 Value Elimination

Like Variable Elimination (Dechter 1999; Zhang & Poole 1994) and Recursive Conditioning (Darwiche 2001), Value Elimination is a query based algorithm for computing posterior probabilities in a Bayesian network. Its input is a Bayesian network containing $n$ discrete valued variables, and $n$ CPTs specifying the probability distribution of each variable given any instantiation of its parents. The Bayes network can be defined by the property that

$$Pr(V_1 = x_1, \ldots, V_n = x_n) = \prod_{i=0}^{n} C_i(x_{i(1)}, \ldots, x_{i(k)}),$$

where $V_i$ is the $i$'th variable, $C_i$ the $i$'th CPT, and $x_{i(j)}$ the value of the $j$'th variable of the $i$'th CPT. That is, the joint distribution over the variables is given by the product of the CPTs.

Value Elimination is a backtracking algorithm that performs a depth-first search in a tree of variable assignments. Hence value elimination is a "conditioning" algorithm. We develop the algorithm in stages.

The first stage is not a backtracking algorithm. Rather it is a simple generate and test (sum) procedure, that searches the entire tree of variable assignments summing the probabilities associated with the leaf nodes.

```
GenAndSum()
1.   V = selectUnAssignedVar()
2.   if V == NONE
3.      prod = 1
4.      foreach CPT c
5.         prod *= eval(c)
6.      return(prod)
7.   sum = 0
8.   foreach d ∈ Dom[V]
9.      assign(V,d)
10.     sum += GenAndSum()
11.     unassign(V)
12.  return(sum)
```

In each recursive call an unassigned variable V is chosen, and each of its values examined (line 8). Visiting a new node in the search tree corresponds to executing line 9, where we make a new assignment. Note that the variable ordering used can be *dynamic*—the recursive calls below each value of $V$ (line 10) might instantiate the remaining variables in a different order. A leaf node is reached when all of the variables have been assigned (line 2), at which point the product of all the CPTs is computed. (`eval` evaluates each CPT with respect to the current variable assignments as set by `assign()`). By the definition of a Bayes network this product is the probability of the current complete assignment. The recursive search below each assignment to $V$ returns the sum of the leaf nodes in the subtree below.

The net result is an exponential computation of one! However, if we have some evidence items $E$ (assignments to some of the variables), and a query variable $Q$, we can make the evidence assignments prior to invoking **GenAndSum**, and select $Q$ as the first variable assigned. The sum returned by **GenAndSum** after $Q$ is assigned the value $d$ will be the probability of $(Q = d) \wedge E$, so the posterior distribution of $Q$ can be obtained by normalizing these probabilities. **GenAndSum** specifies the search tree explored and the semantics of what is being computed. The rest of the development of Value Elimination involves techniques for optimizing the exploration of this tree so that the posterior of $Q$ can be computed without having to visit every node.

The first improvement is to make the procedure into a backtracking search. Backtracking is based on the idea of checking constraints in the tree as soon as all of their variables become assigned and backtracking immediately if a constraint is violated. In the probabilistic context this translates to evaluating the CPTs as soon as they become single valued:

```
Prob-BT()
1.    V = selectUnAssignedVar()
2.    if V == NONE
3.        return (1)
4.    sum = 0
5.    foreach d ∈ Dom[V]
6.        assign(V,d)
7.        prod = 1
8.        foreach CPT c s.t. c is newly single valued
9.            prod *= eval(c)
10.       if prod != 0
11.           prod *= Prob-BT()
12.       sum += prod
13.       unassign(V)
14.   return (sum)
```

A CPT c becomes a single valued function when all of its variables have been instantiated.[1] In **Prob-BT** we accumulate the product of the CPTs that have just become single valued prior to searching the subtree below (line 9). It is not difficult to see that any CPT c that becomes single valued at a node, will appear as a factor in every leaf in the subtree below. Thus early activation of the CPT corresponds to moving common factors outside of summations. Furthermore, if one of the CPTs evaluates to zero we need not search the subtree below (line 10). In **GenAndSum** we would have visited each of the leaves and evaluated its probability (which would have been zero due to the zero common factor). **Prob-BT** can thus save an exponential amount of work over **GenAndSum**, but it must still visit every leaf node that has non-zero probability.[2]

The next improvement to incorporate is intelligent backtracking and nogood recording (Dechter 1990), to obtain further mileage from the zero entries in the CPTs. In the BAYES context a nogood $N$ is a set of variable assignments such that any complete assignment containing $N$ has zero probability. The idea behind nogood learning is to start with sets of variable assignments that force some CPT, $C$, to evaluate to zero. Such sets are nogoods—the joint probability of any complete assignment extending this set will contain a zero factor contributed by $C$. From these *base* nogoods more powerful nogoods can be generated. Specifically, if every possible assignment to a variable $V$ is a member of some nogood, then the union of those nogoods minus all the assignments to $V$ is itself a new nogood (this corresponds to a resolution step). Any complete assignment must make some assignment to $V$. If it also makes all of the assignments in $N$, then it must activate at least one nogood from the set of nogoods that were unioned to form $N$. The new nogood can then participate in the creation of further nogoods.

Nogoods can be unioned together during backtracking search, and then utilized to perform intelligent backtracking, as well as cached to allow the search to avoid future parts of the tree. The methods for accomplishing this are well understood, and are explained in, e.g., (Bacchus 2001). In practice nogood recording typically allows backtracking to save an exponential amount of work. It should also be noted that nogood recording is a much more powerful technique for optimizing zero values during a BAYES computation than the shrink map and zero compression techniques described in (Huang & Darwiche 1996). Nevertheless, it only optimizes the detection of zero probability events; it does not solve the problem of having to visit all assignments having non-zero probability.[3]

The final improvement needed so that backtracking search can achieve good performance on structurally simple networks is a new generalization of the notion of a nogood. This generalization is one of the main original contributions of this paper, and it yields the algorithm we call *Value Elimination*.

---

[1] In the presence of context specific structure the CPT might become single value before all of its variables are instantiated (Section 2.1).

[2] The case where a CPT evaluates to zero is identical to the situation in ordinary constraint satisfaction when a constraint is violated. Hence, additional constraint propagation can be performed to detect other assignments that have zero probability. (Walsh 2002), e.g., presents a backtracking approximation algorithm related to Prob-BT, in which he employs the additional constraint propagation of Forward Checking.

[3] Nogoods for BAYES have been used by (Poole 1996), who presents a tree-search approximation algorithm in which nogoods are generalized to capture events with very low probability.

Nogoods are invariant in the sense that once they are learned they can be used anywhere in the tree. However, they only capture information about factors with zero probability. We develop a similar notion of a "good" that captures information about factors with non-zero probability and that can similarly be used anywhere in the tree.[4]

Returning to **Prob-BT**, consider the recursive call where the *last* uninstantiated variable in the Bayes network, $V$, is selected. (Hence all subsequent recursive calls on line 11 will return 1 via line 3.) Let Dset, the dependency set, be the set of assignments already made higher up in the tree to the variables in the CPTs activated at line 9.[5] Let Sset, the subsumed set, be $\{V\}$ and let Val be the sum over all the values of $V$ computed by the time the procedure reaches line 14.

It is not difficult to see that the computed sum over all of the values of $V$ will always be equal to Val along any branch where $V$ is the last variable instantiated and all of the assignments in Dset have been made—the CPTs activated at line 9 will yield the same values. Hence, along any path that makes all of the assignments in Dset prior to instantiating $V$, we can delay instantiating $V$, and in the penultimate recursive call, after all of the other variables have been instantiated, we can multiply prod by Val rather than making a final recursive call to sum out the values of $V$—that call would have returned Val in any case.

Two important further optimizations can be made. First, if the search is descending down a branch of the tree, we can multiply Val into prod the first time all of the assignments in Dset are made and avoid branching on the variable in Sset in the subtree below—Val is going to be multiplied into every leaf node that lies below. This simply brings the common factor Val up to the highest level of the search tree. Second, when we first compute Val we can immediately pass it back up the tree to multiply it into the prod associated with deepest assignment in Dset, and then avoid branching on $V$ until we backtrack to undo this assignment—this immediately moves the common factor of Val up as far in the tree as possible and avoids having to deal with $V$ while Dset is still active.

Furthermore, we can use these base "goods" to compute more general "goods" or *factors*. A factor $F$ consists of three components, a dependency set $F$.Dset, a subsumed set $F$.Sset, and a value $F$.Val. The semantics of a factor $F$ is as follows. $F$ is **valid** if in the joint distribution when we make all of the assignments in $F$.Dset and sum out all of the variables in $F$.Sset, we will obtain a constant factor equal to $F$.Val times some function of the variables not in

---

[4]The development of a computationally effective notion of a "good" was previously mentioned as an open problem in (Bayardo & Pehoushek 2000).

[5]Ignoring for now the possibility of context specific structure, all CPTs containing $V$ will be activated at line 9.

---

**Value-Elim**(level)
```
1.   V = selectActiveVar()
2.   if V == NONE
3.      return(1)
4.   sum = 0
5.   Dset[level] = {}, Sset[level] = {}
6.   foreach d ∈ Dom[V]
7.      assign(V,d)
8.      markInactiveAtLevel(V,level)
9.      prod[level] = 1
10.     foreach CPT c s.t. c is newly single valued
11.        prod[level] *= eval(c)
12.        Dset[level] ∪= assignments to vars in c
13.     foreach f ∈ FactorCache that is newly activated
14.        prod *= f.Val
15.        Sset[level] ∪= f.Sset
16.        foreach X ∈ f.Sset
17.           markInactiveAtLevel(X,level)
18.        Dset[level] ∪= f.Dset
19.     if prod[level] != 0
20.        Value-Elim(level+1)
21.     sum += prod[level]
22.     unMarkAllInactiveAtLevel(level)
23.  Remove all assignments to V from Dset[level]
24.  Sset[level] ∪= {V}
25.  CacheFactor(Dset[level],Sset[level],sum)
26.  pushBL = Level of deepest assignment in Dset[level]
27.  prod[pushBL] *= sum
28.  Dset[pushBL] ∪= Dset[level]
29.  Sset[pushBL] ∪= Sset[level]
30.  foreach X ∈ Sset[level]
31.     markInactiveAtLevel(X,pushBL)
32.  return
```

Table 1: The Value Elimination Algorithm (without nogood processing).

$F$.Dset ∪ $F$.Sset.

For example, consider a Bayes network defined by the product decomposition $Pr(A)Pr(B|A)Pr(C|B)$, and a factor $F$ with $F$.Dset = $\{B = 0\}$, $F$.Sset = $\{C\}$, and $F$.Val = 1. This factor is valid. In particular, $\sum_C Pr(A)Pr(B = 0|A)Pr(C|B = 0)$ is equal to $xf(A)$ where $x = \sum_C Pr(C|B = 0) = 1 = F$.Val and $f(A) = Pr(A)Pr(B = 0|A)$ is a function of the variables not in $F$.Dset ∪ $F$.Sset

It is not difficult to see that the components Dset, Sset, and Val defined above for the base case of summing out the final variable, form a valid factor according to the above semantics. In particular, every variable in every CPT that $V$ appears in has been instantiated so summing out over $V$ in the joint distribution must produce the claimed constant factor since this is precisely how the factor's value was computed.

To exploit the full power of factors, however, we must develop a method for composing factors into new factors, just as we composed nogoods into new nogoods. Ignoring for now context specific structure in the CPTs, this can be accomplished when the factor is passed up after first being learned. When we branch on a variable we keep a running

Dset and Sset. The contents of these sets are accumulated as we examine the individual values of the variable. Once a factor $F$ is computed, we pass it up to the level of the deepest assignment in $F$.Dset. At that level we multiply the current value's prod by $F$.Val, union $F$.Dset into the variable's running Dset and $F$.Sset into the running Sset. After all of the values of a variable $V$ have been explored, we remove all assignments to $V$ from its running Dset, add $V$ to its running Sset, and create a new factor with this Dset, Sset and the value given by the sum computed at line 14.

PROPOSITION 1 *The new factors computed by the above composition process are valid factors.*

Once factors are computed we can use them in the same way as described above. More precisely, at any node of the search tree were all of the assignments in the Dset of a factor are made and *none* of the variables in the Sset have been assigned[6] we can multiply the current prod by the factor's Val, and avoid branching on any of the variables in Sset in the subtree below.

Adding factor and nogood processing to **Prob-BT** yields the Value Elimination algorithm presented in Table 1. For simplicity, the specification does not include the changes required to implement nogood processing, but these are fairly straight forward.

Value Elimination first checks to see if any cached factors can be activated (line 10). Each activated factor is multiplied into prod[level] As before, the active CPTs are also multiplied into prod[level] (line 11). In both cases Dset[level] is updated to include the assignments that made those factors and CPTs have these specific values (line 12 and 18). Finally, if a zero has not been found the subtree below is searched. On return, prod[level], Dset[level], and Sset[level], would have been properly updated by any relevant factors computed in the subtree below. Finally, when the sum over all the values of $V$ are processed, we create and store a new factor (lines 23–25), compute where to pass it back to (line 26), and update the information at that level (lines 27–30). Finally, since we have added new subsumed variables to the push back level, pushBL, we mark them as inactive until we backtrack to that level (line 31).

### 2.1 Context Specific Structure

Value Elimination can be altered in very simple ways to take advantage of various forms of context specific structure. Here we briefly discuss some of the kinds of structure that value elimination can take advantage of.

---

[6] When dynamic variable orderings are used it could be that we later on instantiate some of the variables in the Sset prior to making all of the assignments in the Dset. In this case the factor value cannot be used, as we are no longer summing over all of the Sset variables.

Local context specific independence (CSI) (Boutilier *et al.* 1996) occurs when a CPT $C$ becomes independent of some of its variables given an instantiation of some of its other variables. For example, if the CPT is a function $C(W, X, Y, Z)$, and we make the assignment $W = 0$, then the new function $C'$ defined by $C'(X, Y, Z) = C(W = 0, X, Y, Z)$ might have the property $\forall y, y'. C'(X, Y = y, Z) = C'(X, Y = y', Z)$. That is, once $W$ is assigned the value 0 the CPT is no longer dependent on the value of $Y$.

To take advantage of the global independencies induced by these local independencies we modify the processing of dependency sets. The modification required is to mark all newly independent variables as being dependent on the assignments that made them independent. In the above example, whenever we subsequently branch on $Y$, we would add $W = 0$ to its final Dset—the value computed when we sum out $Y$ could change if $W$ is not equal to 0 as then $C$ might have an influence on its sum. However, the assignments to $X$ and $Z$ need not appear in $Y$'s Dset—unless these variables influence $Y$ through some CPT other than $C$. This reduction in $Y$'s Dset could also reduce the size of the dependency sets of all of the variables $Y$'s sum ends up being passed up to. Hence, subsumed variables could be pushed back higher in the tree, and the resulting factors could be activated along more different paths. Thus an exponential amount of work could be saved.

Another type of context specific structure discussed in (Boutilier *et al.* 1996) occurs when two values of a variable become equivalent in a certain context. This situation can be detected at the time a variable $V$ is branched on. If there are two values $v$ and $v'$ for $V$ such that two instantiations $V = v$ and $V = v'$ make all of the reduced CPTs that $V$ currently appears in identical, then the subtree below each instantiation would perform the same computations. Hence, we need only explore one of these values assigning the other value the same prod.

It can also be shown that barren variables (Shachter 1986) always yield "null" factors equal to one when summed out. Hence, such variables and their CPTs can be removed from the search without affecting the final answers. Removing one barren variable may in turn make other variables barren, and they can be recursively removed. The end result is identical to Shachter's static barren node removal procedure. However, with CSI it is also possible for variables to become barren dynamically after some assignments are made. Such variables and their CPTs could be removed from the subtree in which they are barren.

### 2.2 Unit Propagation via Forward Checking and Dynamic Variable Orderings

It should be noted that at line 1 Value Elimination is free to choose any variable to instantiate next, i.e., it can uti-lize dynamic variable orderings. In particular, the recursive invocations under the instantiations $V = a$ and $V = b$ are free to choose different variables to instantiate next. Unfortunately, we have not as yet found effective dynamic variable ordering heuristics (in part because we are not yet exploiting context specific structure in our implementation). Nevertheless, we do use dynamic orderings in conjunction with forward checking to realize unit propagation.

Forward checking (Bacchus & Grove 1995) involves examining, at each node of the search tree, all CPTs that are newly reduced to only one uninstantiated variable. If such a CPT evaluates to zero on a particular value of its last uninstantiated variable, say on $V = a$, we know that $V = a$ will contribute zero probability in the subtree below. Forward checking is the process of marking all such zero probability values. If an uninstantiated variable has all of its values marked (perhaps by different CPTs), we exploit our ability to utilize dynamic orderings, and immediately choose that variable to instantiate next. At line 19 `prod[level]` will hence be zero for each value in the variable's domain, a new nogood will be immediately learned, and the search will backtrack. This process allows us to avoid searching in subtrees containing a "deadend variable". Similarly, if an uninstantiated variable has only one unmarked value (thus its value is forced or "unit") it also is immediately chosen next. Thus it is immediately instantiated to its forced value and the consequences of that instantiation forward checked. Preferring forced variables along with forward checking their forced value precisely corresponds to the unit propagation process used in modern SAT solvers. By utilizing both nogood recording and unit propagation, Value Elimination is taking advantage of the key techniques utilized in modern SAT solvers. Thus it is able to get considerable extra mileage out of the zero values in the CPTs.

## 3 Performance Guarantees

As specified Value Elimination is actually a family of algorithms, each member of which is determined by the algorithm used to select the next variable. If we restrict ourselves to static variable selection strategies and ignore any context specific structure then value elimination turns out to be very closely related to two of the fundamental query based algorithms for BAYES: variable elimination and recursive conditioning.

### 3.1 Variable Elimination

Given a ordering of the variables, $\pi = V_{\pi(1)}, V_{\pi(2)}, \ldots, V_{\pi(n)}$, at the $i$'th stage variable elimination sums out $V_{\pi(i)}$ from the joint distribution producing a new function $f_i(X_1, \ldots, X_k)$ over some subset of the variables in the set $\{V_{\pi(i+1)}, \ldots, V_{\pi(n)}\}$. In the absence of context specific structure the following theorem holds.

THEOREM 1 *Let value elimination be run using a* static variable ordering *where at level $j$ variable $V_{\pi(n-j)}$ is branched on.*[7]. *Let $F_i$ be a factor produced by value elimination at line 25 after branching on variable $V_{\pi(i)}$, with $F_i.Dset = \{X_1 = x_1, \ldots, X_k = x_k\}$. Then $F_i.Val = f_i(X_1 = x_1, \ldots, X_k = x_k)$.*

In other words, under a static variable ordering the factors computed are precisely the values of the corresponding function on a particular instantiation. From this it can be shown that the same number of multiplications and summations are required to produce a factor $F_i$ as are required by variable elimination to compute an entry in the table specifying function $f_i$. The only extra work required by value elimination lies in the cost of cache lookup. However, as we will explain below, if we are running value elimination with a static ordering, cache lookup costs can be reduced to the same cost as the array indexing that variable elimination (and recursive conditioning) must use.

COROLLARY 2 *Value elimination when run with the reverse static ordering uses the same time and space as variable elimination.*

Furthermore, nogood recording can allow value elimination to avoid computing large parts of the intermediate functions computed by variable elimination when zero probabilities are present.

### 3.2 Recursive Conditioning

There is also a strong connection between variable elimination and recursive conditioning. Consider first the moral graph associated with the input Bayes network in which each variable is a node and each CPT is a clique over its variables. Instantiating a variable corresponds to deleting the corresponding node in the moral graph along with all of its incident edges. Instantiating a set of variables can thus cut the graph into disjoint components in which the reduced CPTs of each component share no variables with each other.

Value elimination is able to take advantage of components via its tracking of dependency sets. In particular, if at a node in the search tree the moral graph has been divided into $k$ disjoint components by the assignments already made, value elimination will require time proportional to the *sum* of the sizes of these components rather than time proportional the product. That is, value elimination through its use of dependency sets and passing back of values operates as an opportunistic divide and conquer algorithm. Recursive conditioning also utilizes divide and conquer, but the key difference is that it uses a static decomposition scheme, specified by a branch decomposition[8]

---

[7]That is, the last variable eliminated by variable elimination is the first variable branched on by value elimination

[8]This structure is called a d-tree in (Darwiche 2001), but was

The branch decomposition tells recursive conditioning exactly which variables it needs to instantiate at each stage in order to divide the problem into two sub-problems. It then invokes the same procedure on each sub-problem, dividing these into even smaller problems. After the two subproblems have been evaluated it multiplies the results to obtain the answer for the whole problem.

Value elimination, on the other hand, can instantiate the variables according to any strategy, including dynamic strategies. The decompositions that the strategy happens to generate will automatically be detected during the Dset computations on backtrack. Theorem 4 proves that this can yield exponential speedups on some problems. Formalizing these ideas allows the following theorem to be proved:

THEOREM 3 *If recursive conditioning is run with the branch-decomposition (d-tree) B, then from B a static variable ordering strategy can be constructed in linear time under which value elimination will require time and space within a constant factor of the time and space required by the full caching version of recursive conditioning (Darwiche 2001). Furthermore, if we turn off the cache in value elimination, value elimination will achieve the same space-time tradeoff as recursive conditioning without caching (i.e., it will run in linear space and the same order of increase in time).*

It should be noted that these two theorems highlight a close connection between variable elimination and recursive conditioning that was originally made in (Darwiche 2001).

### 3.3 Value Elimination can be Exponentially better

THEOREM 4 *There exists a class of problems on which value elimination using a dynamic variable ordering runs in time $n^{O(\log n)}$ whereas variable elimination, recursive conditioning, and jointree algorithms, require time $O(2^{\sqrt[4]{n}})$ irrespective of the variable ordering (branch-decomposition) they utilize.*

The problems are variants of the string of pearls problem (Bonet *et al.* 1998) originally used to show that ordered resolution (DP) can be exponentially weaker than tree-resolution (DPLL). The proof (Bacchus, Dalmao, & Pitassi 2003) basically shows that although the problem can be solved in quasipolynomial time using a dynamic variable ordering, it requires exponential time for *any* static ordering. Variable elimination, recursive conditioning, and jointree algorithms all utilize static orderings.

Surprisingly, even the simplest version of backtracking upon which value elimination is based, i.e., **Prob-BT**, can achieve this speed up over standard algorithms on these problems.

---

originally called a branch decomposition in the earlier work of (Robertson & Seymour 1991).

### 3.4 Cache Lookup Costs

The substantial difference between the complexity of value elimination and the above two algorithm lies in the time required to do cache lookups. This cost stems solely from our need to support dynamic variable ordering and context specific structure.

In particular, if we restrict value elimination to work without these two features then, as discussed above, the factors computed will simply be instances of the corresponding functions that variable elimination produces. Furthermore, the factors we must multiply together to compute the prod for each value of a variable are also instantiations of functions known prior to search.

Hence, we can allocate tables to store each of the functions that would be produced by variable elimination, and use the Dset of a factor to index and store the factor's value in the table of its associated function. This gives us a fixed address for each factor that could be computed. Since we know the factors we need to compute prod we can "check the cache" by indexing into these tables to see if the factors have already been stored. In other words, cache lookup is reduced to array indexing which is the same as required by value elimination.

In the fully general case, however, cache look up can be a significant overhead. In our implementation this overhead is greatly reduced by utilizing the "watch literal" techniques employed in current SAT solvers (Moskewicz *et al.* 2001), but it remains fairly significant.

Another advantage of value elimination (shared by recursive conditioning (Darwiche & Allen 2002)) is that it can operate in an any space mode. In particular, the cache can be purged at any point in the computation. The remaining computation simply recomputes these purged factors, if in fact it needs them. We have employed a very simple scheme for purging the cache. When we run out of room we remove one half of the stored factors, keeping the half that have the smallest Dset's and largest Sset's (these are more likely to be reused and they required more computation to compute). Many other purging schemes could be investigated.

## 4 Empirical Results

In this section we report on an implementation of value elimination. The implementation includes nogood recording, forward checking, unit propagation via a preference for forced variables, as well as a fully general caching scheme for factors that uses the scheme described above to purge itself when it runs out of memory (the limit was set at 1.5GB).[9] We also perform barren variable removal prior to search (Section 2.1). However, the implementation does not utilize context specific structure. All experiments were run on 2.2GHz Pentium IV machines with 3GB of RAM. The results compare our implementation against Netica, a commercial implementation of the join tree algorithm (Norsys Software Corp.). Comparing these two algorithms is problematic since the join tree algorithm is not query based. Nevertheless, we found that (1) Netica was usually as fast or faster than various implementations of query algorithms we experimented with, (2) it was more robust and better suited for extensive empirical testing, and (3) it is representative of the standard a new algorithm must achieve to be practical.

---

[9]The cache does not employ the static order optimizations described above.

| Network | #Trials | N-Fails | VE-Fails | ≥ 100 | 100–10 | 10–1 | 1–0.1 | 0.1–0.01 | < 0.01 |
|---|---|---|---|---|---|---|---|---|---|
| Water (32) | 414 | 0 | 0 | 371 (0.002s) | 43 (0.033s) | 0 | 0 | 0 | 0 |
| Munin1 (189) | 761 | 0 | 1 | 662 (.043s) | 59 (1.74s) | 30 (21.53s) | 9 (212.5s) | 0 | 1 (7000s) |
| Munin2 (1003) | 48 | 0 | 0 | 0 | 37 (0.06s) | 6 (0.43s) | 5 (2.04s) | 0 | 0 |
| Munin3 (1044) | 16 | 0 | 0 | 0 | 1 (0.09s) | 0 | 13 (2.08s) | 2 (7.29s) | 0 |
| Munin4 (1041) | 621 | 0 | 0 | 0 | 586 (0.12s) | 33 (1.00s) | 2 (9.77s) | 0 | 0 |
| Link (724) | 808 | 799 | 0 | 808 (0.12s) | 0 | 0 | 0 | 0 | 0 |
| Barley (48) | 795 | 0 | 39 | 266 (0.003s) | 22 (0.14s) | 138 (3.22s) | 234 (13.17s) | 72 (61.5s) | 63 (2567s) |

Table 2: Speed up ratio of Value Elimination over Netica on various networks for probabilistic trials. The number of variables in the network is given in brackets after the network name. N-Fails, VE-Fails are the number of trials Netica or value elimination failed on (time out or memory exceeded). The trials are divided into buckets based on speed up ratio (Netica Time/Value Elimination Time). The number of trials in each bucket is given, as well as the average time value elimination requires on trials in that bucket. Failures are placed into the extremal speedup/slowdown buckets.

Most of our experiments involved computing the posterior distribution of a randomly chosen query variable given a randomly chosen evidence item. Before making these random selections, however, we first preprocess the network with a forward checking phase (a polytime computation) to detect variables whose value is forced, and eliminate values with zero probability. The evidence item was then selected at random from the available assignments of an unforced variable and the effect of that evidence was again forward checked. Finally, the query variable was randomly selected from the remaining unforced variables (thus the evidence was not "obviously" contradictory nor was the query variable "obviously" forced by the evidence).

Value elimination was mostly run with a static ordering computed with a min-fill heuristic, but subject to the constraint that the query variable must be branched on first. On some networks, e.g., **Barley**, the static ordering utilized by Netica was more effective. We also experimented with a dynamic ordering based on fill in and cluster size but computed dynamically in the context of the changes made by the previous assignments. The dynamic ordering occasionally produced some good results but was inconsistent.[10]

The **HailFinder** and **Win95pts** networks were found to be simple for both algorithms. We ran 1000 trials on each network with both algorithms completing each trial in less than one second.

The random network **B** (Kozlov) was almost as easy. Running 1000 trials, Netica required less than a second for each trial, whereas value elimination required more than a second on 126 of the trials. In 57 of these trials it was less than 10 times slower than Netica. But on the worst trail it was 75 times slower than Netica, requiring 20 seconds. However, its time on this trial improved to 1.8 seconds when we used our dynamic ordering heuristic.

Data for the more interesting networks **Water, Munin1-4, Link**, and **Barley** is given in Table 2. We ran 1000 trials on each network. Table 2 however excludes those trials on which both algorithms took less than a second, and those where it turned out the evidence was contradictory or the query variable forced. (Polynomial time preprocessing is incapable of detecting all such cases). That is, the table only includes "probabilistic" trials that cannot be solved by purely logical reasoning.

We see that value elimination performs very well, achieving a speed up of more than 100 times on many trials. The failed trial in Munin1 (aborted after 7000 sec.) could be solved in 2057 sec. using Netica's static ordering, during which time the cache was purged 16 times. However, the same trial was solved in only 52.6 sec. when the dynamic ordering was used. Netica took 34.5 sec. to solve this trial. Although Munin2 and 3, are much larger they were quite easy, with almost all of the trials being either easy or solvable by purely logical reasoning. Munin4 was a bit harder, and on this network value elimination was usually 10-100 times faster. Link could rarely be solved by Netica, as for most evidence items it required too much space. Barley was the only network that did not contain many zero entries in its CPTs. Hence, the main advantage of our current implementation, the exploitation of logical reasoning to gain advantage from these zero entries, was not applicable. As a result value elimination could often take a very long time (on 39 trials it timed out after 3600 sec.). How-ever, it still did quite well on many of the trials. We expect that it could do much better once context specific structure is exploited. Finally, on those trials that could be resolved by logical reasoning, i.e., the query was forced or the evidence impossible, value elimination was between 20–2000 times faster than Netica: on these problems value elimination's techniques for logical reasoning achieve their maximum advantage.

---

[10]As mentioned above, dynamic orderings will probably only be truly effective when we are also exploiting context specific structure.

## 5 Conclusions

We have presented an algorithm for BAYES that builds on previous work in backtracking as well as on previous algorithms for BAYES. The algorithm has the advantage of allowing the application of a new set of techniques, like nogood recording, to BAYES, preserving the performance guarantees of standard BAYES algorithms, and having additional flexibility that can allow it to achieve an exponential speedup over previous algorithms on some problems. Empirically, the algorithm displays very promising performance, often being faster than current commercial software. Sometimes, however, it is much slower. Given that the current implementation does not utilize context specific structure nor have very good dynamic heuristics, we feel that these results give strong evidence of the algorithm's considerable potential.